# Approaching the Limits to EFL Writing Enhancement with AI-generated Text and Diverse Learners

*David James Woo*, **Precious Blood Secondary School, Hong Kong**

*Hengky Susanto*, **Department of Science and Environmental Studies, The Education University of Hong Kong, Hong Kong**

*Chi Ho Yeung*, **Department of Science and Environmental Studies, The Education University of Hong Kong, Hong Kong**

*Kai Guo*, **Faculty of Education, The University of Hong Kong, Hong Kong**

**Abstract**

Generative artificial intelligence (AI) chatbots, such as ChatGPT, are reshaping how English as a foreign language (EFL) students write since students can compose texts by integrating their own words with AI-generated text. This study investigated how 59 Hong Kong secondary school students with varying levels of academic achievement interacted with AI-generated text to compose a feature article, exploring whether any interaction patterns benefited the overall quality of the article. Through content analysis, multiple linear regression and cluster analysis, we found the overall number of words -- whether AI- or human-generated -- is the main predictor of writing quality. However, the impact varies by students' competence to write independently, for instance, by using their own words accurately and coherently to compose a text, and to follow specific interaction patterns with AI-generated text. Therefore, although composing texts with human words and AI-generated text may become prevalent in EFL writing classrooms, without educators' careful attention to EFL writing pedagogy and AI literacy, high-achieving students stand to benefit more from using AI-generated text than low-achieving students.

## 1. Introduction

Generative artificial intelligence (AI) technologies have been used in English as a foreign language (EFL) classrooms to enhance students' speaking (Wang, Zou, et al., 2024), reading (Zheng, 2024) and listening (Aryadoust et al., 2024). However, generative AI could especially enhance writing instruction (Guo & Li, 2024). The use of generative AI in the writing classroom has positively influenced students' writing motivation (Huang & Mizumoto, 2024) and writing proficiency (Fathi & Rahimi, 2024). At the same time, using generative AI tools for writing instruction presents challenges (Chen et al., 2024; Woo, Wang, et al., 2024). More deeply understanding students' interactions with generative AI when composing texts (Wang, 2024) could inform more effective pedagogical approaches for composition writing with generative AI. Providing a foreign language (EFL) learning context, this study explores students' interactions with generative AI chatbots in terms of textual contributions that students and chatbots make to co-produce a written composition and the effects of these contributions on the overall quality of the composition.

## 2. Literature review

### *2.1. Applying generative AI in EFL writing classrooms*

A growing body of research suggests the positive impact of AI use on multiple aspects of EFL students' writing skills. For instance, Song and Song (2023) discovered that Chinese EFL students who received AI-assisted instruction showed significant improvements in organization, coherence, grammar, and vocabulary compared to those who received traditional instruction. Similarly, Tsai et al. (2024) confirmed that essays, written by EFL English college students, revised with ChatGPT assistance achieved significantly higher scores, with notable enhancements in vocabulary, grammar, organization, and content.

The integration of AI into writing instruction has also shown benefits on the affective aspects of students' learning experiences. For example, a study by Guo and Li (2024) showed that the use of chatbots positively influenced students' writing motivation, leading to more defined writing goals, increased writing confidence, and a more positive attitude towards writing. Additionally, Teng (2024) found EFL students' perceptions of using ChatGPT had significant positive effects of AI assistance on various aspects of writing, including motivation and self-efficacy.

Other studies showed EFL students encounter challenges when utilizing AI for writing. Chen et al. (2024) attributed students' difficulties in adopting AI-generated corrective feedback on their argumentative essays to overwhelming, generalized, repetitive feedback, perceived lack of utility, stemming from misinterpretations, lack of clear examples, and irrelevant comments. Moreover, Woo, Wang, et al. (2024) also found that, despite students expressing high satisfaction with the learning experience, they reported experiencing high cognitive load during the writing task, particularly during prompt engineering. Thus, identifying students' challenges and needs, and providing corresponding support could further enhance generative AI's positive impact on students' writing.

## 2.2. AI-generated and human-written texts

Educators are worried about generative AI tools' potential to distort students' writing proficiency (Currie, 2023). This distortion could mislead educators to formulate inappropriate teaching strategies and to inaccurately assess student writing. Thus, researchers have sought to differentiate between AI-generated and human-authored essays, finding that AI texts often demonstrate superior quality. For instance, Mizumoto et al. (2024) found significant differences in linguistic features between essays written by Japanese EFL university students and those generated by ChatGPT. Similarly, Nguyen and Barrot (2024) discovered distinct text features that set AI-generated texts apart from human-authored ones. Their findings indicated that essays by a native English lecturer and ChatGPT were rated higher than those by college students and non-native English teachers. Large-scale studies by Herbold et al. (2023) revealed that the argumentative AI-generated essays were rated higher in quality than secondary school student essays. Yang et al. (2024) revealed significant differences in the use of textual, interpersonal, and marked topical themes between texts authored by IELTS teachers and those generated by ChatGPT. Importantly, non-expert readers may struggle to differentiate between AI and human-authored texts. As shown by Porter and Machery (2024), non-experts often misidentified AI-generated poems as human-authored, rating them higher in qualities.

Other studies suggest that AI does not always outperform human authors. Amirjalili et al. (2024) explored the dimensions of authorship and voice in academic writing, focusing on ChatGPT. Their findings revealed the AI's deficiencies in specificity, depth, and accurate source referencing when compared to a university student-authored essay. Similarly, Yang et al. (2024) found redundancy and lack of development in ChatGPT-generated texts. Sardinha (2024) found significant disparities between AI-generated and human-authored texts, and AI-generated texts failed to closely resemble their human counterparts.

The previous studies have approached AI-generated and human-written texts from a binary perspective, as separate entities, but not an integrated one where AI and human writing are interconnected. There has been limited investigation into compositions comprising AI-generated text and human-authored content. Collecting such a body of compositions allows us to explore the fine-grain integration of AI-generated text and a human's own words, and how AI-generated text can complement and enhance the human-authored content, as well as to examine this integration's challenges and implications, including maintaining the ethical use of AI in writing.

### 2.3. Writing with generative AI

The process writing approach conceptualizes writing as a sequence of cognitive stages: planning, drafting, and reviewing (Flower & Hayes, 1981). Though these stages are distinct, writers often move between them in a recursive manner. Research has indicated that teaching students to navigate these stages can lead to more effective writing skill development (Graham & Sandmel, 2011). Furthermore, Crossley et al. (2014) identified distinct profiles of successful students who produced high-quality compositions.

Generative AI tools may shift the knowledge and skills that underpin process writing's stage-specific strategies that guide students towards crafting higher-quality compositions (De La Paz & Graham, 2002). This is because generative AI facilitates a 'machine-in-the-loop' approach to writing, where students collaborate with generative AI technology to complete writing tasks (Gilburt, 2024). This cyclical process involves a student prompting AI, evaluating AI output, and integrating or modifying AI-generated text in a composition. This cyclical process enables AI to contribute to the planning, drafting and reviewing of a composition.

Studies suggest that machine-in-the-loop writing may enhance process writing stages. For example, Wang (2024) found that students used ChatGPT for brainstorming, organizing ideas, and addressing both broad (e.g., argument structure, coherence) and specific (e.g., syntax, diction, grammar) writing issues. Similarly, Guo and Li (2024) found EFL students used self-developed AI chatbots to improve their writing through idea generation, writing outlines, and grammar and spelling checks. Wang, Li, et al. (2024) found postsecondary learners used ChatGPT primarily for brainstorming and seeking inspiration for ideas. Nguyen et al. (2024) found that doctoral students who engaged in iterative, highly interactive processes with a generative AI-powered tool generally achieved superior writing task performance, while those who merely used generative AI as an additional information source and maintained a linear writing approach tended to have lower performance.

Nonetheless, our understanding of students' drafting and revising strategies when writing with a machine-in-the-loop remains limited, rarely applied to compositions generated by students along with AI output. A notable exception is the study by Woo, Susanto, et al. (2024), which identified distinct profiles of students who produced high-quality compositions by editing AI-generated and human-written text in specific ways. However, their study's contributions to understanding how AI contributes to the fine-grain drafting and reviewing of compositions were limited by the use of small, open-source language models such as GPT-2, GPT-NEO-1.3B and GPT-J-6B.

This study investigates the linguistic characteristics of AI-generated text in student compositions, identifying patterns of interaction with AI-generated text, and identifying how different interaction patterns may correspond to varying types of learners' writing. Because new and prolific state-of-the-art large language models may shift how students' draft and revise when writing with a machine-in-the-loop, there is potential to identify new patterns of interaction with AI-generated text, and profiles of students who write and edit high-quality compositions using the latest generative AI tools, not merely ChatGPT. Besides, we may find profiles of students who write and edit lower-quality compositions using the same tools. The results can provide insight into using generative AI to enhance the writing process and inform EFL writing instruction. The research questions (RQs) guiding this study are:

- *RQ1*: What are the language features of AI-generated text in students' compositions written with AI?
- *RQ2*: What interaction patterns with AI-generated text can be identified in students' compositions and how do they differ?
- *RQ3*: How do interaction patterns with AI-generated text influence and potentially benefit different types of learners' writing?

## 3. Methodology

### 3.1. Research context and participants

59 EFL students participated and came from seven Hong Kong secondary schools that represent different levels of academic achievement. Students attended a two-hour workshop at their school. The workshop's title was 'How to attempt a writing task with ChatGPT support' and Appendix 1 presents the workshop learning design. The first author led the workshop. In the workshop, students learned an explicit EFL writing approach, either genre-based or process-based. Students then learned prompt engineering to support that approach to writing. Next, students performed a feature article writing task (see Figure 1). The specific task was chosen by each school's teacher in charge and was taken from a Hong Kong Diploma of Secondary Education (HKDSE) examination writing paper that all Hong Kong mainstream school students must take in secondary six. Their compositions could not exceed 500 words.

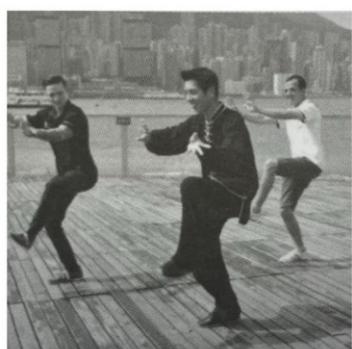
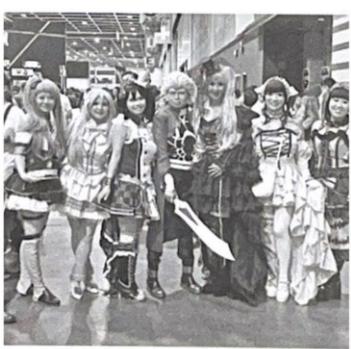
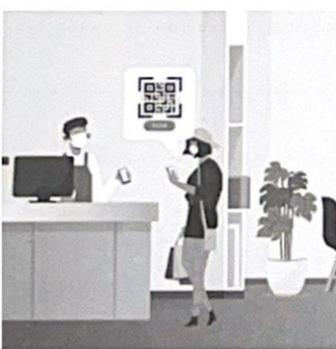

**Figure 1**. The three possible writing tasks

Students were instructed to use their own words and words from at least one generative AI chatbot in their written composition. Students could use as many or as few of their own words or chatbot words as necessary. Although students were introduced to the POE app and its collection of state-of-the-art chatbots in the workshop (see Figure 2), students could use any chatbot software. They could use as many chatbots as necessary. Students attempted the task on an iPad, laptop or desktop computer.

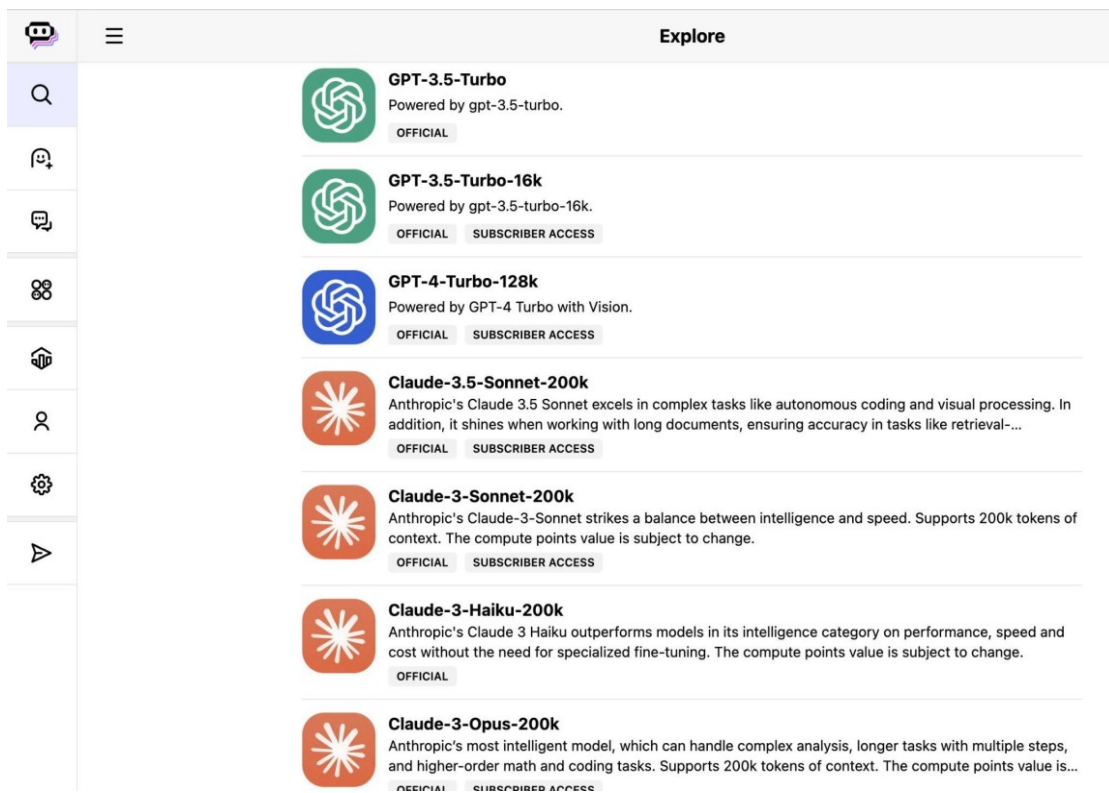

**Figure 2.** A screenshot of the POE app featuring different chatbots

Students were required to self-report AI-generated text in their compositions by highlighting that text according to the chatbots they used and a color key (see Figure 3). Students colored their own words black.

- Highlight each company's chatbots' words in a different color:
    - Anthropic in ORANGE: Claude etc.
    - Google in GREEN: Gemini; Google-Palm
    - Meta in BLUE: Llama etc.
    - OpenAI in RED: GPT-4; ChatGPT; GPT-3.5 etc.
    - Other in PURPLE: Assistant; Mistral; Mixtral; Solar; StableDiffusionXL; Web-Search

**Figure 3.** The color key for highlighting AI-generated text

Students were informed that their compositions would be scored. Before scoring, the compositions would be rendered in black color to remove any indicators of AI-generated text. Student identifiers would also be removed from texts. The first author and the English teacher would then award scores for each composition's content (C), language (L) and organization (O) according to the HKDSE writing paper marking scheme, a common marking scheme for English language texts in Hong Kong schools (see Appendix 2). To improve reliability, the first author and the English teacher attended a standardization meeting before scoring, and in case of discrepancies, averaged their scores.

### 3.2. Data collection

The deadline to submit the task was set for two weeks after a workshop. Students wrote their texts on Google Docs and shared these with the first author. Compositions exceeding 500 words were not collected for analysis.

### 3.3. Data analysis

We performed a content analysis (Neuendorf, 2017) to identify language features of AI-generated text in students' compositions. We adopted measures per

Woo, Susanto, et al.'s (2024) study, operationalizing language features as a composition's basic structure and organization, and the AI-generated text's syntactic complexity. First, we counted a composition's number of words including the number of words written by a student and the number of AI-generated words. Then we counted the number of AI-generated chunks or text instances and the word length of each chunk, and the number of human chunks. Each AI-generated chunk was categorized into one of three production unit lengths (Lu, 2010) with the sentence as the essential unit (Hyland, 2003): (1) a short AI chunk is AI-generated text less than five words in length; (2) a medium chunk is sentence length or at least five words in length; and (3) a long chunk is longer than a sentence. We composed descriptive statistics for compositions' basic structure and organization, and the AI-generated texts' syntactic complexity. Additionally, we examine these statistics alongside the academic profiles of students' schools and human-rated scores.

To identify interaction patterns with AI-generated text, we used the multiple linear regression (MLR) (Aiken et al., 2005) statistical technique. MLR describes the relationship between different variables, taking into account the effect of each variable. As we are interested in how AI-generated text and a student's text contribute to the overall quality of a composition, our MLR explores the relation between C, L, O scores as well as the total score of a student's composition with the composition's number of AI-generated and human chunks and words. The specific MLR equation for examination is presented in the Results section.

To explore how interaction patterns with AI-generated text can benefit different types of learners' writing, we utilized cluster analysis—an unsupervised machine-learning technique that groups data points based on their similarities (Dempster et al., 1977). By applying this technique, we can statistically identify inherent patterns in our data by grouping students with similar characteristics into distinct clusters. Specifically, we investigated how the syntactic unit length of AI-generated text chunks, and the number of AI-generated words incorporated into a student's composition would impact the student's C, L, O scores and the total score. The specific algorithm for cluster analysis is presented in the Results section.

**4. Results**

*4.1. What are the language features of AI-generated text in students' compositions written with AI? (RQ1)*

Table 1 presents the basic structure of 59 compositions and the syntactic complexity of their AI-generated text. The average composition length was 443.44 words, with AI contributing 281.10 words (63.39%). A composition averaged 8.22 AI chunks with an average chunk length of 57.04 words. Long AI chunks appeared in 52 compositions, medium in 45, and short in 31.

**Table 1**. The utilization of chunks and words by students in their compositions

| No. | Type of Chunk | No. of Texts with this Type | Avg. No. Words (Std. Dev.) | Avg. Count (Std. Dev.) | Avg. Length (Std. Dev.) |
|---|---|---|---|---|---|
| 1 | All | 59 | 443.44 (80.19) | | |
| 1.1 | Human | 59 | 162.34 (149.87) | 7.51 (6.04) | |
| 1.2 | AI | 59 | 281.10 (150.44) | 8.22 (6.52) | 57.04 (54.83) |
| 1.2.1 | Long AI | 52 | 254.29 (153.45) | 3.6 (2.17) | |

| | | | | |
|---|---|---|---|---|
| 1.2.2 | Medium AI | 45 | 61.51 (53.55) | 3.78 (3.13) |
| 1.2.3 | Short AI | 31 | 9.45 (10.76) | 3.94 (5.63) |

To explore the large standard deviation of AI words in compositions, Figure 4 shows the distribution of the percentage of AI words. The figure shows a wide range of AI word utilization from 0-10% AI words (n=2) to 90-100% AI words (n=17). Compositions most frequently comprised 90-100% AI words and the majority of compositions consisted of at least 70% AI words (n=33).

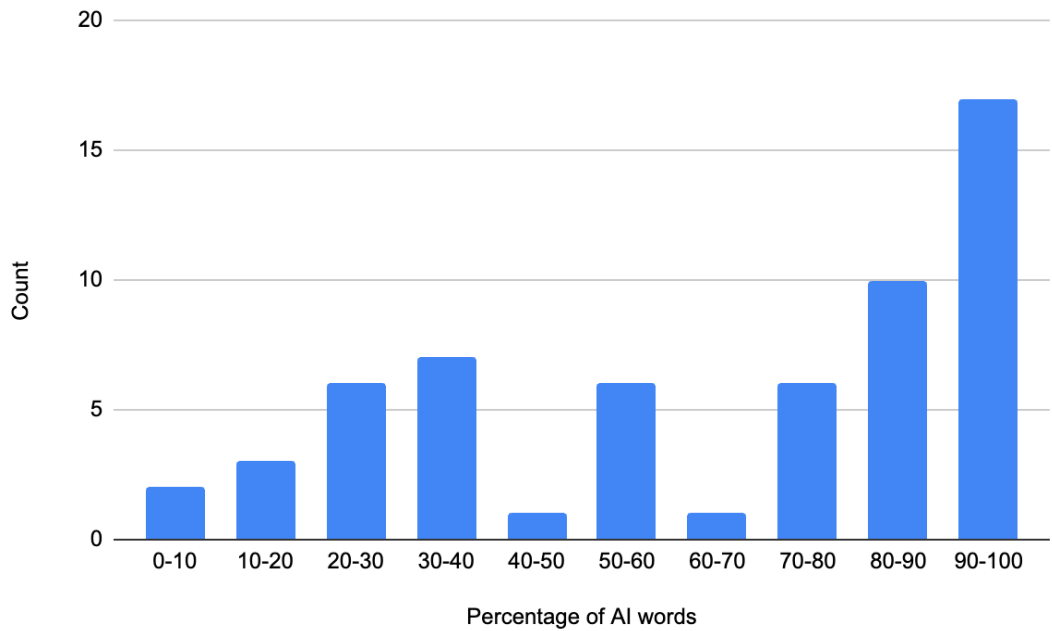

**Figure 4**. The Distribution of the Percentage of AI Words among the 59 Compositions

Table 2 presents composition structure information organized by school. We composed two school bands, a higher band comprising students with academic achievement percentiles from 55 to 99, and a lower band comprising students from 1 to 44. There were 37 compositions from high-banding school students, and 22 from low-banding students. The average percentage of AI words in a school's compositions ranged from MCW's 34% to BHSS's 85%. The average number of AI chunks ranged from LSTWCM's 4.86 to MCW's 12.75.

**Table 2**. The utilization of AI-generated words by school

| Academic Band | Academic Percentiles of Students | School Name | N | Avg. No. Words (Std. Dev.) | Avg. % of AI Words (Std. Dev.) | Avg. No. AI Chunks (Std. Dev.) |
|---|---|---|---|---|---|---|
| Low | 1-11 | LSTWCM | 7 | 439.14 (87.31) | 0.79 (0.25) | 4.86 (2.27) |
| Low | 22-44* | BHSS | 7 | 455.71 (41.06) | 0.85 (0.23) | 7 (4.86) |
| Low | 22-44* | PBSS | 2 | 336.5 (47.38) | 0.67 (0.27) | 7.5 (0.71) |
| Low | 33-44* | IKTMC | 6 | 467.5 (53.38) | 0.67 (0.37) | 7.5 (6.66) |
| High | 55-77* | MCW | 4 | 432.5 (83.08) | 0.34 (0.27) | 12.75 (11.56) |
| High | 77-99* | SPSLT | 11 | 404.09 (115.01) | 0.66 (0.28) | 8.64 (9.84) |
| High | 88-99 | HPCCSS | 22 | 465.73 (66.08) | 0.58 (0.31) | 8.91 (4.93) |

Table 3 presents the human-rated composition scores. The highest possible total score was 21, and the average total score was 14.65 (3.28 standard deviation). Comparing the C, L and O criteria, students on average scored the highest for the L criterion and the lowest for O. Notably, L and O scores could not exceed the content score by one mark plus or minus.

**Table 3**. Composition scoring results

| Scoring Item | Full Score | Average Score | Standard Deviation |
|---|---|---|---|
| Content | 7 | 4.8 | 1.27 |
| Language | 7 | 5.25 | 1.02 |
| Organization | 7 | 4.61 | 1.1 |
| Total | 21 | 14.65 | 3.28 |

Table 4 presents average composition scores organized by school bands. Except for students from IKTMC, low-banding students wrote compositions with a lower average total score than high-banding students. In the following sections, we explore associations between interaction patterns with AI-generated text and students' school banding.

**Table 4**. Banding of students by school academic profile

| Academic Band | Academic Percentiles of Students | School Name | N | Average Total Score (Std. Dev.) |
|---|---|---|---|---|
| Low | 22-44* | PBSS | 2 | 12.25 |
| Low | 22-44* | BHSS | 7 | 12.64 |
| Low | 1-11 | LSTWCM | 7 | 13.86 |
| High | 55-77* | MCW | 4 | 13.88 |
| High | 77-99* | SPSLT | 11 | 13.89 |
| Low | 33-44* | IKTMC | 6 | 15 |
| High | 88-99 | HPCCSS | 22 | 16.18 |

***4.2. What interaction patterns with AI-generated text can be identified in students' compositions and how do they differ? (RQ2)***

The MLR model we examined is given by Equation (1) where $x_1$ = number of short AI chunks, $x_2$ = number of medium AI chunks, $x_3$ = number of long AI chunks, $x_4$ = number of human chunks, $x_5$ = number of AI words, and $x_6$ = number of human words.

$$Score = m_1 x_1 + m_2 x_2 + m_3 x_3 + m_4 x_4 + m_5 x_5 + m_6 x_6 + C, \quad \text{Equation (1)}$$

By the method of least squares (Dekking et al., 2005), we could obtain the best fitted parameters $m_i$ in the MLR model given by Equation (1). First, we fitted the C, L, O and the total scores of all the 59 compositions by using the number of AI-generated and human chunks and words in the compositions. After that, we examine the correlations between these scores and variables for all 59 compositions, as illustrated in Table 5.

As shown in the last column of Table 5, although we only examined MLR models which were linear, the models with the best fitted parameters explained the different C, L, O as well as the total scores for all the 59 submissions, with small p-values, i.e., a high statistical significance. Nevertheless, the values of $m_i$ did not fully represent the correlation between the scores and the variable $x_i$, since $x_i$ was not normalized, for $i = 1,2,\ldots,6$. We thus computed the so-called partial correlations (Brown & Hendrix, 2005), which are correlations between the score and the variables after removing the co-dependence on all other variables in Equation (1). For instance, the partial correlation $\tilde{C}(score, x_1)$ between the score and $x_1$ is related to the fitted parameter $m_1$ through the following relation:

$$\tilde{C}(score, x_1) = m_1 \times \frac{residual(x_1)_{x_1 = f(x_2,\ldots,x_6)}}{residual(score)_{score = f(x_2,\ldots,x_6)}} \quad \text{Equation (2)}$$

where $residual(x_1)_{x_1=f(x_2,…,x_6)}$ and $residual(score)_{score=f(x_2,…,x_6)}$ correspond to the residuals of the MLR models in fitting $x_1$ and the score respectively and separately, using only the factors $x_2, …, x_6$ (Dekking et al., 2005). Hence the co-dependence on variables $x_2, …, x_6$ are eliminated in the partial correlation $\tilde{C}(score, x_1)$.

Table 5. The partial correlations between various scores and variables for all 59 compositions

|  | No. short AI chunks | No. med. AI chunks | No. long AI chunks | No. of human chunks | No. of AI words | No. of Human words | Model p-value |
|---|---|---|---|---|---|---|---|
| Content (C) score | -0.169 | 0.099 | -0.063 | 0.099 | 0.387** | 0.389** | 3.63 x $10^{-3}$ |
| Language (L) score | -0.087 | 0.150 | 0.022 | -0.004 | 0.434** | 0.437** | 2.73 x $10^{-3}$ |
| Organization (O) score | -0.125 | 0.098 | -0.100 | 0.037 | 0.392** | 0.409** | 1.19 x $10^{-2}$ |
| Total score | -0.135 | 0.119 | -0.052 | 0.050 | 0.417** | 0.424** | 3.62 x $10^{-3}$ |

*Note.* A pound (#), a single star (*) and a double star (**) correspond to the cases of statistical significance with a p-value less than 0.1, 0.05 and 0.01 respectively. The last column shows the p-value of the corresponding linear regression model in explaining the various scores across the 6 variables.

To understand how AI-generated text contributes to students' compositions, we first examined the partial correlations between the scores and the number of chunk types and word types from all 59 submissions, as shown in Table 5. The partial correlations between all the scores, including the total scores, and the number of AI and human words were positive with a value around $\tilde{C} \approx 0.4$ and were highly statistically significant with a p-value less than 0.01. All other partial correlations between the scores and the number of AI and human chunks were smaller in values, with both positive and negative values, and were not statistically significant. These results imply that if we consider all 59 compositions regardless of school banding, the scores were mainly dependent on the number of words in the submissions, but not on how the AI-generated chunks and words were incorporated or distributed. Specifically, the more words in the student compositions, regardless of AI-generated or human-written, the higher the C, L, O and total scores.

To explore how AI-generated texts contribute to compositions from more competent and less competent writers, we analyzed the MLR models and the partial correlations between the scores and the number of words and chunks based on the banding of the students' school. Table 6 shows the partial correlations and model p-values obtained from the 37 submissions in the high-banding schools, which were generally consistent with those obtained from all the 59 submissions in Table 5. Nevertheless, we found that the values of the partial correlations between the number of AI and human words in Table 6 are higher than those in Table 5, implying that the scores are even more strongly dependent on the number of words in the submissions from the high-banding school students. These results imply that the phenomenon of more competent writers submitting longer submissions is more prominent in high-banding schools. Such conjecture is also supported by the model p-values, as they are smaller in Table 6 than Table 5.

Table 6. The partial correlations between various scores and variables for the 37 compositions from the high-banding schools

|  | No. short AI chunks | No. med. AI chunks | No. long AI chunks | No. of human chunks | No. of AI words | No. of Human words | Model p-value |
|---|---|---|---|---|---|---|---|
| Content (C) score | -0.385# | -0.077 | -0.172 | 0.333# | 0.518** | 0.492** | $2.19 \times 10^{-3}$ |
| Language (L) score | -0.312# | -0.003 | -0.178 | 0.230# | 0.654** | 0.552** | $3.62 \times 10^{-4}$ |
| Organization (O) score | -0.381# | -0.122 | -0.151 | 0.292 | 0.496** | 0.528** | $6.57 \times 10^{-3}$ |
| Total score | -0.383# | -0.076 | -0.176 | 0.308 | 0.578** | 0.544** | $1.14 \times 10^{-3}$ |

*Note.* A pound (#), a single star (*) and a double star (**) correspond to the cases of statistical significance with a p-value less than 0.1, 0.05 and 0.01 respectively. The last column shows the p-value of the corresponding linear regression model in explaining the various scores across the 6 variables.

For compositions from high-banding schools, we observed a slight statistical significance represented by a p-value between 0.05 and 0.1 (indicated by a "#" sign in Table 6), for the *negative* partial correlations between the C, L, O and the total scores with the number of short AI chunks, as well as a *positive* partial correlation between the C and L scores with the number of human chunks. The former may indicate an ineffective drafting and revising strategy adopted by the students from high banding schools and the latter may indicate that these same students are capable of writing on their own, independent of AI-generated text.

Table 7 presents the partial correlations between the scores and the number of AI and human words and chunks and model p-values for compositions from low-banding schools. Compared to those from high-banding schools in Table 6, the model p-values are much larger than 0.05, implying a low explanatory power of the MLR models on the scores, unlike the very high explanatory power of the MLR models for high-banding schools. Besides, there is a lack of statistically significant partial correlation between the scores and the number of AI words, unlike those statistically significant and high $\tilde{c}$ observed for high-banding schools. Moreover, the partial correlations between the scores and the number of human words are only slightly statistically significant. These results imply that more AI words are not helping students from low-banding schools to get a higher score for their compositions, since the students may not be proficient in drafting with and revising AI-generated text.

**Table 7**. The partial correlations between various scores and variables for the 22 compositions from the low-banding schools

|  | No. short AI chunks | No. med. AI chunks | No. long AI chunks | No. of human chunks | No. of AI words | No. of Human words | Model p-value |
|---|---|---|---|---|---|---|---|
| Content (C) score | 0.342 | 0.376 | 0.386# | -0.400# | 0.198 | 0.402# | 0.507 |
| Language (L) score | 0.463 | 0.473 | 0.528 | -0.522 | 0.131 | 0.466 | 0.230 |
| Organization (O) score | 0.339# | 0.354# | 0.295* | -0.393* | 0.276 | 0.429# | 0.513 |
| Total score | 0.387 | 0.407 | 0.413 | -0.445 | 0.210 | 0.439# | 0.421 |

*Note.* A pound (#), a single star (*) and a double star (**) correspond to the cases of statistical significance with a p-value less than 0.1, 0.05 and 0.01 respectively. The last column shows the p-value of the corresponding linear regression model in explaining the various scores across the 6 variables.

We observed a statistically significant and *negative* partial correlation between the content (C) and the organization (O) scores and the number of human chunks in Table 7, unlike the positive partial correlations observed in Table 6. Furthermore, there is a statistically significant and *positive* partial correlation between the C score and the number of long AI chunks, as well as the O scores with all the number of short, med and long AI chunks. These results suggest that from low-banding schools, compositions largely comprising long AI-generated chunks are likely to get a high C and O scores, and human revision negatively impacts the scores. That would indicate that students from low-banding schools are less competent not only at drafting and revising a composition in their own words but also at integrating human and AI writing. Therefore, less competent writers relying on verbatim output from generative AI chatbots may lead to higher scores, but their attempts to revise that AI-generated text may lower the scores.

*4.3. How do interaction patterns with AI-generated text influence and potentially benefit different types of learners' writing? (RQ3)*

For cluster analysis, we employed the Expectation-Maximization (EM) Algorithm (Dempster et al., 1977), which utilizes Gaussian distribution to probabilistically estimate the likelihood of data points belonging to a particular cluster.

Our cluster analysis considered eleven features described in Table 8. However, based on our observations, inputting all the features into the algorithm could lead to information saturation, making analysis difficult. To address this issue, we categorized the features into five sets: Principal Features and Supplementary 1 to 5. The Principal set was paired with the features from one of the Supplementary sets. Thus, in different experiments, the algorithm included features from the Principal and one of the Supplementary sets.

**Table 8.** Cluster analysis features

| Features Description | Features set |
|---|---|
| The percentage of AI-generated words used in a story (the entire article) | Principal |
| The percentage of words written by human in a story | Principal |
| The student's C score, the L score, the O score, and the average total score | Principal |
| The total number of words generated by AI in a story | Supplementary 1 |
| The number of human chunks | Supplementary 1 |
| The total number of words generated by human in a story | Supplementary 1 |
| Total number of LLM chatbot used in the process of writing the story | Supplementary 2 |
| Number of specific LLM chatbot used in the writing (Claude, Google_Palm, GPT-X, and others) | Supplementary 2 |
| The number of AI chunks (short, medium, and long) | Supplementary 3 |
| Total number of words in each of the AI chunk category (short, medium, and long) | Supplementary 4 |
| Average number of words in each AI chunk in all categories (short, medium, and long) | Supplementary 5 |

The total number of clusters utilized in the analysis ranges from 2 to 4, for *k = 2, 3, 4*. In other words, the 59 submissions were clustered into *k* groups. When *k > 4*, the results would be a breakdown of patterns similar to those observed when *k = 4*. Thus, higher *k* would result in diminishing return, where more details would not necessarily lead to new understanding. For these reasons, we presented only the results with cluster size of *k = 2, 3, 4*.

*4.3.1. The AI-generated texts utilization pattern in higher and lower school banding (k=2)*

This setting allowed us to analyze the interaction patterns of AI-generated text between high- and low-banding schools. Additionally, high- and low-banding schools were indexed as 1 and 0, respectively, which we used to estimate the population of students from these two profiles using the following equation.

$$school\ banding\ index\ = \frac{1}{|s|}\sum banding\ index, for\ s \in Cluster\ i,$$

*Equation (3)*

where s and |s| is submissions and number of submissions in Cluster $i$, for $i = \{0, 1\}$. Thus, *school banding index* closer to 1 indicates there were more students from high-banding schools in the cluster and vice versa.

Table 9 presents results where the algorithm used features from either the Principal and Supplementary 1 sets as inputs. Importantly Cluster 1 has a school banding index of 0.8, and appears less reliant on AI-generated text in composition writing, indicating that students from high-banding schools are more capable of writing in their own words, independent of generative AI chatbot use.

**Table 9.** Outputs generated by using features from either the Principal and Supplementary 1

| Cluster ID | % of AI words | % of human words | School Banding index |
|---|---|---|---|
| 0 | 71 | 29 | 0.53 |
| 1 | 53 | 47 | 0.8 |

*4.3.2. The relationship between AI-generated texts distribution and total score (k = 3)*

In this experiment with a cluster size of k = 3, the algorithm used features from the Principal and Supplementary 1 as inputs to generate the clusters. Table 10 presents the distribution of AI-generated words and human words, with each cluster's results summarized as average values. Furthermore, we normalized the total scores to a range between 0 and 10 to easily compare. Compared to other clusters, Cluster 0 shows the highest average total score and the largest percentage of human words, suggesting these students are more capable of drafting and revising a composition independent of AI-generated text. Cluster 1 shows a high average total score and a large percentage of AI words, suggesting these students are capable of beneficially drafting and revising a composition with AI-generated text. On the other hand, Cluster 2 shows the lowest average total score and these may have adopted the wrong strategy, being less capable of either writing without AI-generated text or integrating AI-generated text in a composition.

**Table 10.** Outputs generated by using features from the Principal and Supplementary 1

| Cluster ID | Avg. total score | % of AI words | % of human words | No. human words | No. of AI words |
|---|---|---|---|---|---|
| 0 | 8.1 | 35.9 | 64.1 | 328.68 | 147.18 |
| 1 | 7.3528 | 90.99 | 9.012 | 43.8 | 428.72 |
| 2 | 5.4 | 54.6 | 45.3 | 179.1111 | 195.1111 |

*4.3.3. Student tendency in utilizing AI (k = 4)*

For this experiment, we set the number of clusters to k = 4 and the algorithm used features from the Principal and Supplementary 1 sets as inputs. Figure 5 is a stacked bar chart and presents student submissions clustered into four groups based on the percentage of AI-generated words and the average total score. The students could be categorized into four profiles: 1) students with low scores but a high percentage of AI-generated words; 2) students with high total scores but fewer AI-generated words; 3) students with low scores and fewer AI-generated words; and 4) students with high scores but a high percentage of AI-generated words. The profiles show that higher utilization of AI words can benefit some students' scores (Cluster 3) but not others (Cluster 0). At the same time, higher utilization of human words can benefit some students' scores (Cluster 1) but not others (Cluster 2).

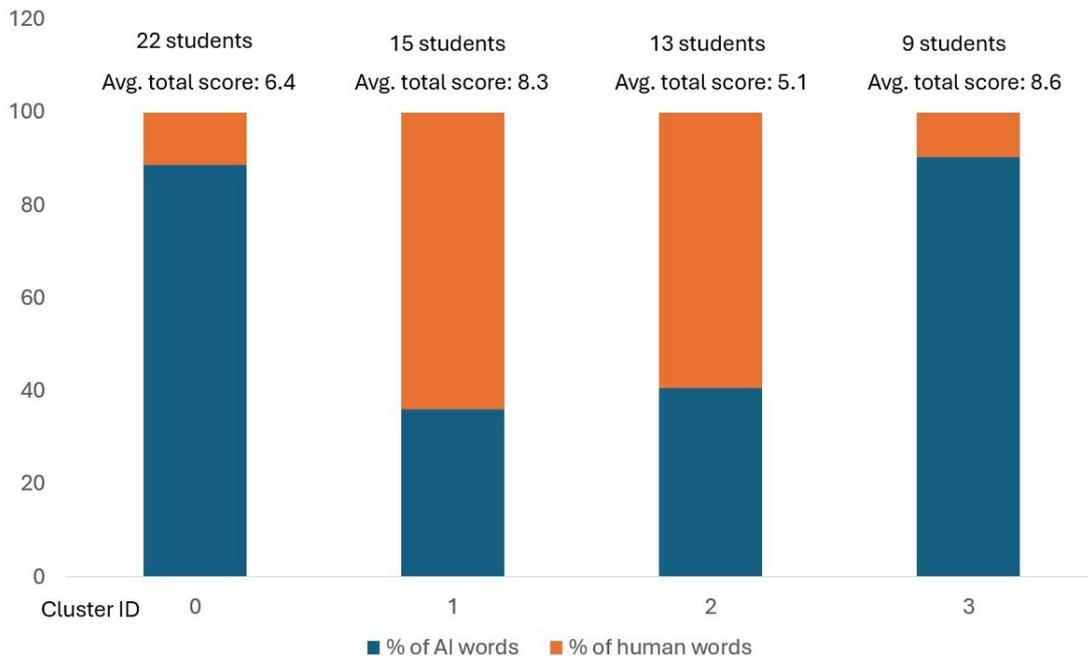

**Figure 5.** Distribution of the percentage of AI and human generated words

Figure 6 presents AI-word utilization organized by school academic profile. The distribution of students from high-banding schools amongst clusters does not show a strong preference for these students relying heavily on AI-generated text or their own words. Specifically, in Clusters 0 and 3, there are 11 and 6 students, respectively, who use more AI-generated words in their writing, while their peers in Clusters 1 and 2 consist of 12 and 8 students who use fewer AI-generated words. In contrast, most students from low-banding schools rely heavily on AI-generated text.

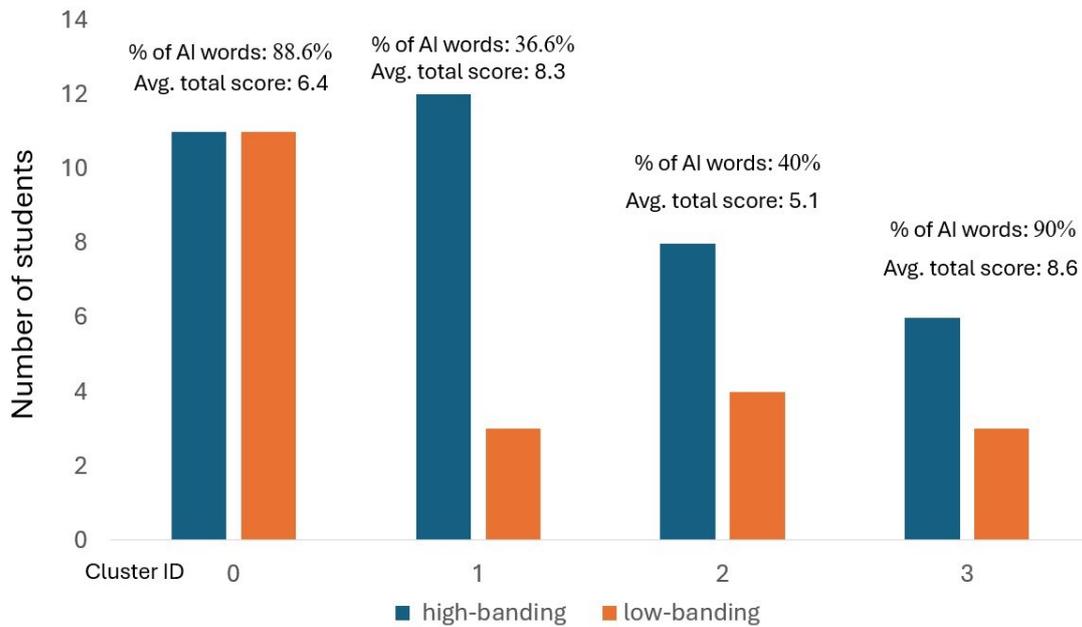

**Figure 6.** Student distribution based on school banding

*4.3.4. Relationship between AI generated texts and CLO scores (k = 4)*

Table 11 presents the results when the algorithm used features from the Principal and Supplementary 4 sets and split the students into high- and low-banding schools. We observe interaction patterns with varying word lengths of short-, medium- and long-AI generated chunks do not necessarily lead to higher total scores, especially evidenced in cluster 0. For example, high-banding students in Clusters 0, 1, and 2 use more words per short AI chunk than their low-banding peers. Similarly, in Clusters 0, 1, 2, and 3, high-banding students also use more words per medium AI chunk. For long AI chunks, high-banding and low-banding students in Clusters 1, 2, and 3 have similar word lengths. However, in Cluster 0, low-banding students with the lowest total scores write significantly more words per long AI chunk than high-banding students. Ultimately, neither low-banding nor high-banding students may be more proficient at strategizing their use of AI in their compositions with varying word lengths of AI-generated syntactic units to improve overall quality of their compositions.

Table 11. Words per AI chunk and total score by school banding

| School Banding | Cluster ID | % of AI words | % of human words | words per short AI chunk | words per med AI chunk | words per long AI chunk | Avg. total score |
|---|---|---|---|---|---|---|---|
| Low | 0 | 67.28 | 32.7 | 0.75 | 8 | 95 | 3.8 |
| high | 0 | 31.8 | 68.2 | 2.5 | 12 | 36.25 | 4.88 |
| Low | 1 | 51.1 | 48.8 | 1.6 | 14 | 28 | 8.41 |
| high | 1 | 36.7 | 63.3 | 2.77 | 20.4 | 28.3 | 8.28 |
| Low | 2 | 69.46 | 30.5 | 1.2 | 10.2 | 46.2 | 6 |
| high | 2 | 66.3 | 33.7 | 1.9 | 17.6 | 46.5 | 6.5 |
| Low | 3 | 91.76 | 8.2 | 0.3 | 4.3 | 142.8 | 7.2 |
| high | 3 | 90.2 | 9.8 | 0.2 | 11.4 | 142.5 | 7.75 |

## 5. Discussion

### 5.1. Major findings

Unlike studies that had sought differences in essays written by humans and those by generative AI (Mizumoto et al., 2024; Nguyen & Barrot, 2024), our study explored 59 compositions written with AI-generated text and a person's own words, with the person retaining full control of the writing process. Compared to Woo, Susanto, et al.'s (2024) study of EFL students' human-AI compositions, our study shows the basic organization and structure of a typical human-AI composition written by an EFL student has shifted to greater reliance on AI-generated text in terms of higher average number of AI words and chunks, higher average word count and higher percentage of AI words. Students are taking advantage of advances in AI chatbot capabilities, including longer context lengths (Munkhdalai et al., 2024) which facilitate lengthier outputs of human-like text (Brown et al., 2020) -- and reasoning abilities that improve output's coherence.

However, when considering the average total composition score of 14.65 out of 21 marks, students taking advantage of chatbots' capabilities to generate lengthy texts does not lead to the highest quality writing. This could be attributed to some EFL students lacking sufficient traditional literacy, for example, to compose a 500-word article, and to evaluate lengthy AI-generated output. Besides, some EFL students may lack familiarity with the feature article genre (Hyland, 2003), impacting the planning, drafting and revising of this text type. In terms of writing with a machine-in-the-loop, students may lack prompt engineering skills (Zamfirescu-Pereira et al., 2023) for appropriate and detailed AI-generated content that supports their planning, drafting and revising.

Results from the MLR analysis identified an essential, statistically significant interaction pattern: composition scores are mainly dependent on the number of words in a composition, but not on how AI-generated text is integrated whether by syntactic units, number of AI-generated text chunks or chunk word length. Furthermore, we organized our analysis by school academic profile and found interaction pattern differences between students from high-banding schools and those from low-banding schools. Because students from high-banding schools appear capable of writing a feature article independent of AI-generated text, they have more strategies to integrate AI-generated text at their disposal, for example, to rely more or less on it. Although a particular strategy did not appear particularly beneficial for scores, ultimately these high-banding students benefit their scores by being able to write up to 500 words. In contrast, using more AI words did not help students from low-banding schools to achieve higher scores. At the same time, human revision of AI-generated text negatively impacted their scores. EFL students from low-banding schools appear especially weak at composing feature articles. Their most beneficial writing strategy could be drafting a composition completely with AI-generated text.

Results from the cluster analysis identified interaction pattern differences between high-banding and low-banding school students. For example, in the clustering with k=2, students in the higher school banding index of 0.8 appeared less dependent on AI-generated text in composition writing and more capable of writing in their own words. In the clustering with k=4, high-banding students did not appear to show a strong preference for relying either on AI-generated text or their own words but most students from low-banding schools relied heavily on AI-generated text. The cluster analysis results also identified interaction patterns of distinct learner profiles apart from school banding. In the clustering with k=3, we identified groups who achieved different overall scores and who adopted distinct distributions of AI-generated words and human words, with the highest scoring group showing the largest percentage of human words. At k=4, we identified four distinct learner profiles: one with high AI-word use and high total score; a second with high-AI word use and low total score; a third with low-AI word use and high total score; and a fourth with low AI-word use and low total score.

Ultimately, our cluster analysis shows that the same interaction pattern with AI-generated text can benefit some students' writing but not others. At the same time, these patterns appear unequally distributed amongst students from different academic profiles, as high-banding students more strategically used a variety of interaction patterns to draft and revise a composition with AI-generated text. Therefore, it appears EFL students from low-banding schools may encounter more challenges when utiliziing AI for writing, perhaps experiencing difficulty navigating through the cognitive stages of the writing process (Flower & Hayes, 1981) and suffering high cognitive load during the writing task (Woo, Wang et al., 2024). Moreover, since low-banding students heavily rely on AI-generated text, there is more potential that generative AI tools will distort these students' writing proficiency (Currie, 2023), whereas high-banding students may use the tools to enhance their existing writing proficiency.

*5.2. Implications*

Our study has provided empirical evidence of how AI-generated output from writing with a machine-in-the-loop contributes to EFL students' drafting and reviewing of a composition. Furthermore, we advanced understanding of a typical composition written with AI-generated text and human words. We contribute distinct profiles of successful students who produced high-quality compositions (Crossley et al., 2014) from the context of from high- and low-academic achievement Hong Kong secondary schools. Our learner profiles expand on those found in Woo, Susanto, et al.'s (2024) study and correspond with advances in generative AI capabilities.

Practically, our study shows limits to AI-generated text's enhancement of EFL writing, particularly through the lens of academic achievement. Although low-banding students are taking advantage of much AI-generated text to increase the volume of words in a written composition, these students might lack process writing skills and sufficient traditional literacy to avail themselves of other strategies to plan, draft and to revise with AI-generated text, and to enhance AI-generated text in a composition with their own words. Thus, educators should first aim to develop these students' process writing capacity, evidenced by students' drafting accurate and coherent compositions in their own words. For high-banding students who may have sufficient capacity to draft compositions in their own words, educators could focus on finer-grain skills, such as students transitioning from using long to short AI chunks when composing written work with AI-generated text. Besides, since composition quality may be constrained by students' capabilities to effectively plan and revise a composition, and to prompt generative AI, educators should develop high-banding students' strategies to select generative AI chatbots and to effectively prompt them to support the planning and revision of a composition. Ultimately, educators are crucial for developing all EFL students' capabilities as independent writers, who can use generative AI to enhance writing, not to replace and distort human effort.

### *5.3. Limitations and future directions*

Our study has focused on the drafting and reviewing stages of process writing. Besides, it has examined machine-in-the-loop writing through a corpus of compositions written with human words and AI-generated text. Future studies can consider exploring how EFL students plan their compositions when writing with a machine-in-the-loop. They could also explore other phases of writing with a machine-in-the-loop, such as prompt engineering, evaluating AI-generated output and editing human words and AI-generated text. To do so, researchers should employ different methods such as recording EFL students' screens and using think-aloud protocols. Researchers could identify practices that are beneficial for human-rated scores. Comparative studies could compare human-authored essays, AI-authored essays and human-AI-authored essays for language features and other qualities. They could also compare results between our Hong Kong EFL secondary school students and another sample of EFL secondary school students.

## 6. Conclusion

This study has demonstrated that EFL secondary school students' interactions with AI-generated text can enhance their composition writing. However, the extent of the benefit depends on specific interaction patterns and students' existing writing competence. EFL students from low academic achievement schools may face the most difficulty writing a composition without relying heavily on AI-generated text. Compared to students from high academic achievement schools, they are most limited when interacting with AI-generated text so as to maximize potential benefits to their writing quality. Although machine-in-the-loop writing is becoming increasingly pervasive and presents opportunities for curriculum redesign in the writing classroom, without educators' attention to EFL writing pedagogy and AI literacy, some schools and their students stand to benefit more than others.

**ACKNOWLEDGEMENT** The data used in this study was processed by *Yilin Huang* from The Education University of Hong Kong, Hong Kong, China.

**Declaration of Generative AI and AI-assisted technologies in the writing process**

During the preparation of this work the authors used ChatGPT in order to improve readability and language. After using this tool, the authors reviewed and edited the content as needed and take full responsibility for the content of the publication.

**Appendix 1**. Workshop learning design

| | |
|---|---|
| Title | How to attempt a writing task with ChatGPT support |
| Time | 2 hours |
| Purpose | To develop Hong Kong students' and teachers' competence to use ChatGPT for English writing enhancement |
| Intended learning outcomes (LT) | 1. I can understand genre / process and its approach to writing.<br>2. I can understand ChatGPT and identify its tasks<br>3. I can understand prompts and identify their categories<br>4. I can write prompts for different writing stages<br>5. I can independently develop a text with the support of ChatGPT |
| Learning activities (ILOs) (minutes) | 1. Pre-workshop questionnaire (5 minutes)<br>2. Introduction to writing approach (10 minutes)<br>3. Introduction to AI, chatbots and ChatGPT (5 minutes)<br>4. Model prompt types with examples (25 minutes)<br>5. Guided practice applying prompts to writing stages for an HKDSE task (25 minutes)<br>6. Introduction to writing task and setting up (10 minutes)<br>7. Independent practice on HKDSE writing task (30 minutes)<br>8. Wrapping up and post-workshop questionnaire (10 minutes) |
| Materials (written language) | 1. Generative AI tools on POE app on iPads<br>2. Google Docs<br>3. Shared Google Drive folder:<br>a. Contest website (English language)<br>b. Marking scheme (English language) |

|  | c. Pre- and post-workshop questionnaires (English and Chinese languages) |
|  | d. Workshop slidedeck (English language) |
|  | e. Worksheets (English language) |
|  | 4. iPads / desktops |
|  | 5. Poll Everywhere (English language) |
| Instructional language | English |

**Appendix 2.** HKDSE English Language Paper Two (Writing) Marking Scheme

| Marks | Content (C) | Language (L) | Organization (O) |
|---|---|---|---|
| 7 | · Content entirely fulfills the requirements of the question<br>· Totally relevant<br>· All ideas are well developed/supported<br>· Creativity and imagination are shown when appropriate<br>· Shows a high awareness of audience | · Very wide range of accurate sentence structures, with a good grasp of more complex structures<br>· Grammar accurate with only very minor slips<br>· Vocabulary well-chosen and often used appropriately to express subtleties of meaning<br>· Spelling and punctuation are almost entirely correct<br>· Register, tone and style are entirely appropriate to the genre and text-type | · Text is organized extremely effectively, with logical development of ideas<br>· Cohesion in most parts of the text is very clear<br>· Cohesive ties throughout the text are sophisticated<br>· Overall structure is coherent, extremely sophisticated and entirely appropriate to the genre and text-type |
| 6 | · Content fulfills the requirements of the question<br>· Almost totally relevant<br>· Most ideas are well developed/supported<br>· Creativity and imagination are shown when appropriate<br>· Shows general awareness of audience | · Wide range of accurate sentence structures with a good grasp of simple and complex sentences<br>· Grammar mainly accurate with occasional common errors that do not affect overall clarity<br>· Vocabulary is wide, with many examples of more sophisticated lexis<br>· Spelling and punctuation are mostly correct | · Text is organized effectively, with logical development of ideas<br>· Cohesion in most parts of the text is clear<br>· Strong cohesive ties throughout the text<br>· Overall structure is coherent, sophisticated and appropriate to the genre and text-type |
| | | · Register, tone and style are appropriate to the genre and text-type | |
| 5 | · Content addresses the requirements of the question adequately<br>· Mostly relevant<br>· Some ideas are well developed/supported<br>· Creativity and imagination are shown in most parts when appropriate<br>· Shows some awareness of audience | · A range of accurate sentence structures with some attempts to use more complex sentences<br>· Grammatical errors occur in more complex structures but overall clarity not affected<br>· Vocabulary is moderately wide and used appropriately<br>· Spelling and punctuation are sufficiently accurate to convey meaning<br>· Register, tone and style are mostly appropriate to the genre and text-type | · Text is mostly organized effectively, with logical development of ideas<br>· Cohesion in most parts of the text is clear<br>· Sound cohesive ties throughout the text<br>· Overall structure is coherent and appropriate to the genre and text-type |
| 4 | · Content just satisfies the requirements of the question<br>· Relevant ideas but may show some gaps or redundant information<br>· Some ideas but not well developed<br>· Some evidence of creativity and imagination<br>· Shows occasional awareness of audience | · Simple sentences are generally accurately constructed.<br>· Occasional attempts are made to use more complex sentences.<br>· Structures used tend to be repetitive in nature<br>· Grammatical errors sometimes affect meaning<br>· Common vocabulary is generally appropriate | · Parts of the text have clearly defined topics<br>· Cohesion in some parts of the text is clear<br>· Some cohesive ties in some parts of the text<br>· Overall structure is mostly coherent and appropriate to the genre and text-type |

|   |   |   | · Most common words are spelt correctly, with basic punctuation being accurate<br>· There is some evidence of register, tone and style appropriate to the genre and text-type |   |
|---|---|---|---|---|
| 3 |   | · Content partially satisfies the requirements of the question<br>· Some relevant ideas but there are gaps in candidates' understanding of the topic<br>· Ideas not developed, with possible repetition<br>· Does not orient reader effectively to the topic | · Short simple sentences are generally accurate.<br>· Only scattered attempts at longer, more complex sentences<br>· Grammatical errors often affect meaning<br>· Simple vocabulary is appropriate<br>· Spelling of common words is correct, with basic punctuation mostly accurate | · Parts of the text are generally defined<br>· Some simple cohesive ties used in some parts of the text but cohesion is sometimes fuzzy<br>· A limited range of cohesive devices are used appropriately |
| 2 |   | · Content shows very limited attempts to fulfil the requirements of the question<br>· Intermittently relevant<br>· Some ideas but few are developed<br>· Ideas may include misconception of the task or some inaccurate information | · Some short simple sentences accurately structured<br>· Grammatical errors frequently obscure meaning<br>· Very simple vocabulary of limited range often based on the prompt(s) | · Parts of the text reflect some attempts to organize topics<br>· Some use of cohesive devices to link ideas |

| | | | |
|---|---|---|---|
| | · Very limited awareness of audience | · A few words are spelt correctly with basic punctuation being occasionally accurate | |
| 1 | · Content inadequate and heavily based on the task prompt(s)<br>· A few ideas but none developed<br>· Some points/ ideas are copied from the task prompt or the reading texts<br>· Almost total lack of awareness of audience | · Multiple errors in sentence structures, spelling and/or word usage, which make understanding impossible | · Some attempt to organize the text<br>· Very limited use of cohesive devices to link ideas |
| 0 | · Totally inadequate<br>· Totally irrelevant or memorized<br>· All ideas are copied from the task prompt or the reading texts<br>· No awareness of audience | · Not enough language to assess | · Mainly disconnected words, short note-like phrases or incomplete sentences<br>· Cohesive devices almost entirely absent |

Notes:

The rubric descriptors for content, language and organization are taken from HKDSE English writing rubric descriptors.

Content mark cannot exceed 1 if the text type is not complete.

Creativity in content refers to the details, transformation and originality of ideas

Language and organization marks cannot exceed +/- 1 of the content mark.